# 3D Object Quality Prediction for Metal Jet Printer with Multimodal Thermal Encoder Network

Rachel (Lei) Chen
*HP INC*
Palo Alto, CA, USA
lei.chen1@hp.com

Wenjia Zheng
*HP INC*
Seattle, WA, USA
wenjiaz06@gmail.com

Sandeep Jalui
*University of Florida*
Gainesville, FL, USA
sandeepjalui5@gmail.com

Pavan Suri
*HP INC*
Corvallis, OR, USA
pavan.suri@hp.com

Jun Zeng
*HP INC*
Palo Alto, CA, USA
jun.zeng@hp.com

*Abstract—* **With the advancements in 3D printing technologies, it is extremely important that the quality of 3D printed objects, measured in metrics including mechanical properties, and dimensional accuracies should meet the customer's specifications. Various factors during Metal printing affect the printed part's quality, including the powder quality, the printing stage parameters, the print part's location inside the print bed, the curing stage parameters, and the metal sintering process. With the large data gathered from HP's MetJet printing processes, AI techniques can be used to analyze, learn, and effectively infer the printed part quality metrics, as well as assist in improving the print yield. In-situ thermal sensing data captured by printer-installed thermal sensors such as the Heimann camera contains the part thermal signature of fusing layers. Such part thermal signature contains a convoluted impact from various factors that occur during metal printing. In this paper, we use a multimodal thermal encoder network to fuse the data of different nature including the video data and vectorized printer control data, and extract part thermal signatures with a trained encoder-decoder module. The pre-trained model with efficient thermal feature extraction is then fused with printer control parameters for downstream tasks including part dimensional accuracy prediction and part porosity prediction. We explored the data fusing techniques and stages for data fusing, the optimized end-to-end model architecture indicates an improved part quality prediction accuracy.**

## I. INTRODUCTION

HP is a world-leader in additive manufacturing with a portfolio of products covering polymer and metals. HP's Metal Jet 3D printing system S100 uses precision control of heat to trigger the phase transition of metal powder material to create metal parts of shape and strength as designed.

Hp's Metal Jet prints green parts (which refers to the part generated through an additive manufacturing process after initial material formation, but before subsequent processes are performed to the part). Green parts are placed in the sintering oven to generate the final metal piece. Each print stage introduces prospective variables to the final part quality, including the powder initial quality assessment, powder baking, loading mechanics, binder setting, printing, thermal control, curing, and sintering. Final part geometrical accuracy and consistency remains the top challenge to manufacturing yield of the Metal Jet printed parts. "Green part" porosity [1] is the top root cause for this part inconsistency. Green parts out of additive manufacturing processes could be more porous than other traditional technologies (e.g., MIM), and the green parts after sintering could result in 25% to 50% volumetric shrinkage [2]. Yet there are limited solutions for quantitate assessment or the prediction of green part porosity for a specific printer, a specific print design and powder batch in metallic additive manufacturing. There is no or limited solution to predict the green part porosity and subsequently infer the best design solution to optimizing the print bucket yield. For a given printed piece, manually measuring the critical dimensional metrics, and labeling the visual defect can be one approach to assess the part quality. However, when onboarding a new application, using a new powder batch, or stabilizing a new hardware system, running manual, expensive tunning cycle to achieve the objective green part quality is required.

HP is developing Digital Twin to virtualize the complex material phase transitions during production [3] such that the user can predict and then optimize both the design parameters and the process control parameters to improve part quality and ultimately manufacturing yield. As part of the effort in the HP Digital Twin, in-situ thermal data is analyzed and applied for part quality assessment. Thermal signature, which can be collected with in-situ sensors, is an important factor that convolves a mixed impact of process control parameters and correlates to the 3D printed part quality. In this work our contributions include:

1. Proposed a data pipeline to efficiently extract part-level thermal signature with an encoder-decoder network. The well-trained encoder extracts a thermal latent vector to represent the part thermal signature for a new print design layout.
2. Applied the proposed data pipeline to extract part-level geometry information and validated the reconstruction result, which enables subsequent green part quality prediction capability before print.
3. Proposed a model architecture to integrate the encoded thermal signature (before print) or geometry information (after print) to green quality prediction model, validated the green part quality metrics that is applicable to a specific printer, a specific print design, a powder batch, and the part location inside the print bed.

II. RELATED WORK

*A. Additive manufacuring of metal*

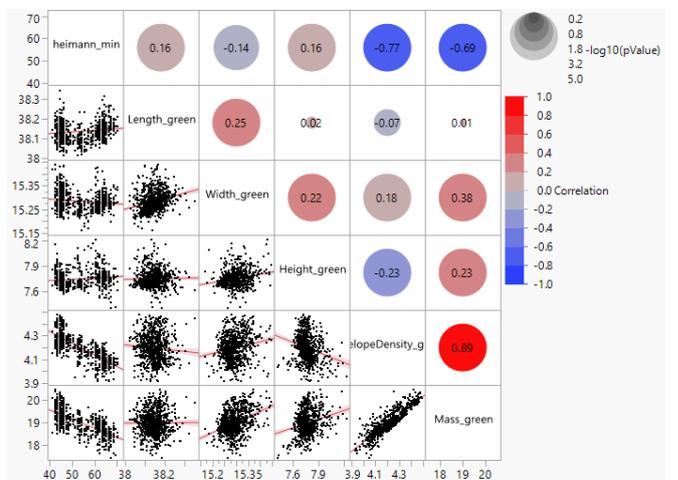

Fig. 1. Correlation analysis on the part-level thermal impact towards the green quality dimension and part density. Part level thermal aggregated minimum temperature shows high correlation to green part density.

Additive manufacturing technology of metal materials like stainless steel, and copper has applications in a wide range including conductive components, machinery, medical implants, automotive, etc. [1][4] The primary challenge of metal printing with AM, such as Binder jetting (BJ) [5], Selective Laser Melting (SLM) [6], Selective electron beam melting (SEBM) [7] is producing fully dense, homogeneous parts. For example, the Binder Jetting printed copper parts can have a porosity range from 2.7% to 16.4%, leading to varied part properties including thermal conductivity, and electrical conductivity. Other part qualities that need to be closely controlled include part dimension and mechanical properties such as tensile strength. HP's Metal Jet S100 printer is commercially available for mass production of high-quality industrial parts, providing end-to-end supply chain solutions that are customer-centric with high repeatability. Part quality data are collected in-house for optimizing overall manufacturing yield, which is a key step towards commercializing. In AM processes in general, the effect of specific process parameters' quantitative correlation to part quality are to be established yet necessary in determining the optimal production printers' process parameter values[10][11], quality control and acceptance criteria are not well standardized and may vary from different customer applications [8][9], some examples may include the variations in the series of a printer, the mean accuracy, and bucket variation in part dimension, the density of bucket after printer calibration, other criteria also include energy saving, cost.

*B. Digital Twin for Industry 4.0*

The emerging technology concept Digital Twin, also known as digital avatars or digital masters, serves as a virtual replica of the real physical asset or system. It is a computational model built upon the collected physical data based on physical theories and realizes the interactions between the digital world and real world. A Digital Twin system produces simulations or forecasts the future of the physical counterpart [12][13][14], through the rapid (real-time) collection of online data and information analytics, enables the engagement and feedbacks with the physical environment [15][16][17]. The adoption of Digital Twins in Industry 4.0 has proven advances including cost reduction, efficiency improvement and production consistency etc., its wide range of applications covers from supply chain management [24], smart logistics and different manufacturing processes including product and process design, optimization, quality control etc.

Digital Twin in additive manufacturing assists in achieving the higher-quality production of parts with the integrated predictive powder of simulation and in-situ sensor gathered data [18]. One example of the in-situ sensor is the thermal imaging systems that have been long serving the monitoring and improvements of the AM processes [19][20][21]. HP Multi Jet Fusion's closed loop thermal control system measures temperatures at fine resolution on the material bed [22] and monitors the regional, as well as layer-to-layer energy receival, allowing the process control such as fusing and cooling, and defects control including thermal bleed.

*C. Multimodal deep learning*

Multimodal deep learning (MMDL) [31][32] describes deep learning models that take a variety of data modalities such as linguistic, acoustic, and visual data [24][25]. With the nature of heterogeneity of data sources in many real-world application domains that contain multisensory data collection, such as in the healthcare field [26][33][37], sequencing prediction [38], review analysis [36], urban infrastructure planning [36][39], multimodal machine learning is gaining attention yet comes with challenges in the interconnections between different data modalities, representation, alignment, and reasoning. Model architectures need to be specially designed to outperform the uni-modal networks [27][34]. MultiBench [28] provides a large-scale benchmark for multimodal learning that spans 10 modalities and 10 prediction tasks. Different data fusing techniques, as well as stages of fusing, are studied exhaustively in [29][30].

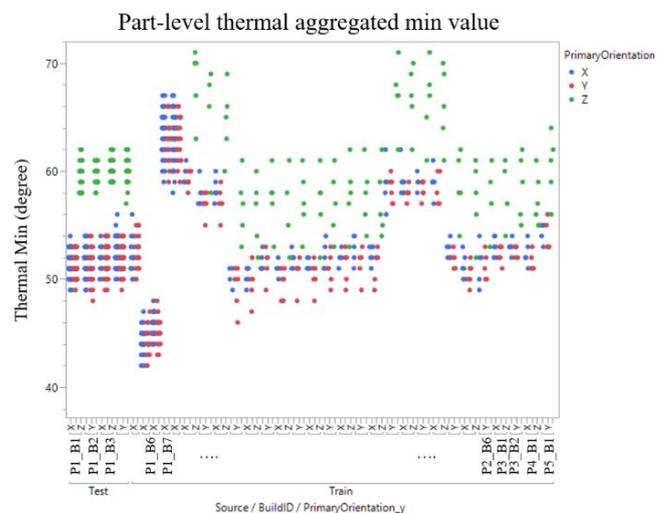

Fig. 2. Part-level thermal minimum temperature distribution across the train and test set, across the different printers and print builds, as well as the print orientations. "P" indicates the printer number and "B" indicates the build ID.

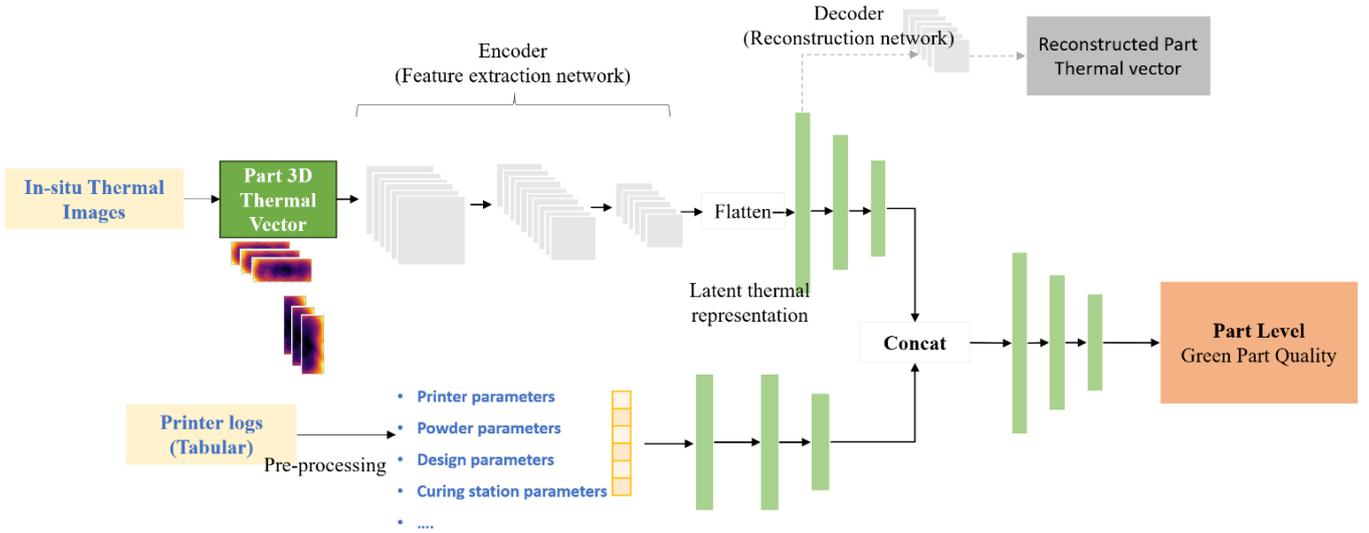

Fig. 3. Proposed Multimodal thermal encoder network architecture.

III. THE PROPOSED METHOD

To predict the green porosity for a given printed piece in an automated fashion, or to enable the capability of prediction before the actual print run in order to realize the subsequent tasks such as the design solution optimization and print process parameters searching for a specific printer, a specific print design and powder batch, we proposed a network architecture to leverage the data collected from a modality of sources – the printer logs as well as the in-situ measurement. The printer logs contain various printer control parameter settings, and in-situ sensors' real-time print history data are time series, recorded for every print layer. We pulled these modalities of data from the HP S100 printers' database. Specifically, the printer logs contain information including, but not limited to the printer controllable parameters, the powder lot properties, the build file design parameters, applied binder level, print part orientations, and other parameters such as the curing station applied temperatures; the in-situ thermal sensing data can be collected with defined framerate to indicate the thermal history for each print fusing layer at the pixel level.

*A. Data preprocessing*

- TRS bars: Following the "Standard Test Methods for Notched Bar Testing of Metallic Materials" [40][41], we used the specimen geometry of same design size rectangular bars of in our experiments to validate the feasibility of the proposed architecture with fast data collection. We collected and experimented with the full-solid specimen data printed with both 316 and 17-4PH stainless steel. The similar 3D geometry design is used in the standard test method for the transverse rapture strength of powder metallurgy specimens; thus, we refer to the tested 3D design geometry as TRS bars in the following description.

- Printer logs (or telemetry data) are collected in the tabular format, we preprocessed with data cleaning, aligning, and filtering out the desired features. We collected training data from 29 printed builds (from 5 different printers of the same firmware version to factor in the printer-to-printer variabilities in the training set) that contain various geometries and have TRS bars inserted among the parts in the print bed as proxy. The training dataset contains different print control parameters such as the binder amount applied, varied shadowing strategy (as a design variable) on parts, print layer thickness, powder batch and powder recycle runs etc. After preprocessing, for each part we were able to collect a feature size of 25~30 as a one-dimensional vector. The entire training dataset after data preprocessing contains 761 TRS bar parts from different print bed location, different print orientations and different printers. Each TRS bar's measured green part quality is then aligned and collected, including but not limited to the dimensional measurement length, width, height, as well as mechanical properties represented by part envelope density. We adopt the four most frequently used properties in the following experiments.

- The in-situ thermal data is collected at print-time for each print run, measuring the pixel-level fusing layer thermal distribution. We took one image per fusing layer at the pre-defined framerate, to exclude the measurement difference introduced at different fusing state. We then preprocess the image data by applying un-distortion, data filtering and cropping to map out the clean region. We cropped out each part's interest (ROI), which is a 3d-vector by aligning the part positioning description. With this data preprocessing pipeline, we were able to obtain the in-situ thermal vector for each part as a 3-dimensional vector. For the TRS bars printed at different orientations (vertical or horizontal placement orientation), we transposed and reshaped the 3D vector to be of the same vector shape of 18 (width) x 35 (length) x 7 (height).

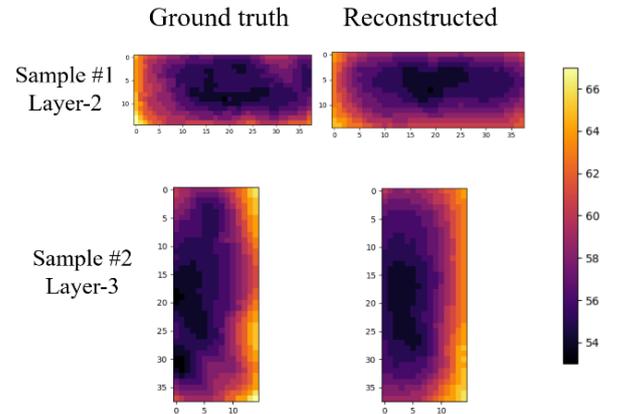

Fig. 4. Reconstruction examples from two different printers and print locations and orientations, each was taken at a sample z-height layer.

- For the last data processing step, we mapped the part 3-dimensional thermal vector data with the part tabular feature data to form the complete data instance to train the proposed multimodal network. We also computed the part-level aggregated thermal feature for analysis purpose. As shown in Fig. 2, we show the part-level thermal minimum temperature distribution across the train and test set across the different printers and print builds, as well as the print orientations. During model training, each data instance is provided with its measured green part quality vector represented by the TRS bar dimension and envelope density to minimize the predicted green part quality difference.

### B. Model architeccture and training

Our proposed model architecture is as shown in Fig.3. Part thermal vector is inputted to an encoder for thermal signature extraction. Then both the thermal latent representation vector and the printer parameter vector are passed through several dense layers before concatenating and passing through the final fully connected layers to predict the part quality vector. Fig. 3 shows this end-to-end training pipeline to predict the part quality, as well as the reconstruction module to validate the latent representation extraction feasibility.

For the encoding the thermal latent representation and reconstruction (decode), we first experimented with implementing the stand-alone encoding and decoder pipeline to validate the reconstruction capability. We experimented with encoding and reconstructing the part thermal vector with both an Autoencoder-based network, and a 3D Variational Autoencoder-based network to compare the performance quantitatively. With the same training and testing data, we were able to achieve much better reconstruction accuracy with the 3D-VAE-based reconstruction model. We further adopted the 3D-VAE-based architecture in the Multimodal thermal encoder network, adding the decoder module and its reconstruction loss to the main objective function.

### C. Objective functions

For the reconstruction loss function design in the 3D-VAE-

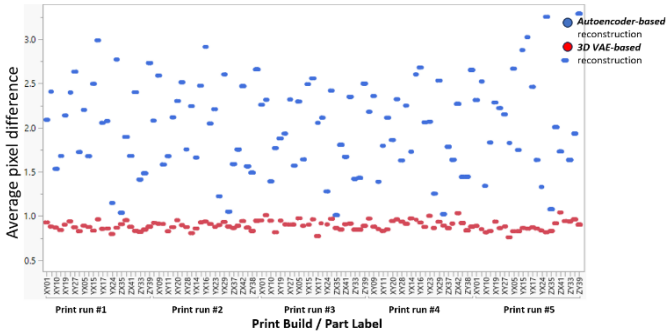

Fig. 5. Model reconstruction accuracy comparing two encoder-decoder model architectures, measured in Average pixel difference (ADP).

TABLE I. MODEL VERSIONS & DIMENSIONAL PREDICTION ACCURACY COMPARISON

| Model version | % error | | | (s.t.d.) | | |
|---|---|---|---|---|---|---|
| | *L* | *W* | *H* | *L* | *W* | *H* |
| Without thermal input | 0.65 | 0.39 | 1.50 | 0.33 | 0.04 | 0.09 |
| With thermal sequential input | 0.37 | 0.25 | 1.07 | 0.15 | 0.03 | 0.06 |
| With thermal latent vector | **0.15** | **0.13** | **0.34** | **0.13** | **0.02** | **0.03** |

based model, we primarily focused on adjusting the weight for the Mean Squared Error (MSE) loss to get faithful reconstructed images on pixel-basis. The Kullback-Leibler Divergence (KLD) loss, which acts as a regularization term, is more related to the latent space's structure and the generation of novel samples, thus is added as secondary loss term. As defined below, $X$ refers to the inputs (both thermal vector and the tabular parameter data), $Y$ refers to the part green quality vector, $p_i$ refers to the $i$-th part data instance. The complete objective function consists of the green part quality prediction mean square error (MSE), and the weighted term of part thermal reconstruction loss and KLD loss.

$$L(Y|X) = \sum((\hat{p}_i - p_i)^2 + w1 * (\widehat{Re}_{p_i} - T_{p_i})^2 + w2 * KLD)$$

## IV. EXPERIMENTAL RESULTS

Our experiment data and the model prediction accuracy were collected and demonstrated with HP S100 on testing builds of tensile bars with stainless steel.

### A. Reconstructoin quality

For the experiment on training the stand-alone encoder and reconstruction module, we used a training data size collected from 31 print builds across 5 printers of the same series. We randomly separated the data into 819 parts for training and 91 parts for validation. With each part's thermal trace, after passing through the pre-processing data pipeline, each TRS part has a thermal input vector size of 18x35x7 pixels, referring details to Section 3A. We trained both an Autoencoder-based network, and a 3D Variational Autoencoder-based network, then tested the well-trained models on 3 new print builds covering 2 different printers, 131 data instances in total. The reconstruction quality on part thermal vector comparing the two model architectures measured in part average pixel difference (ADP) is shown in Fig. 5. The Autoencoder-based reconstruction model has an ADP of 1 to 3 degrees. In contrast, the 3D-VAE-based reconstruction model shows much-improved prediction accuracy, achieving an average of around or below 1 degree ADP. We thus adopted the 3D-VAE-based model architecture to integrate into the Multimodal thermal encoder network to compute the reconstruction loss and green quality prediction. Fig. 4. Shows two examples of two tested TRS parts' reconstruction result comparing to the ground truth, each taken at a different z-height slice.

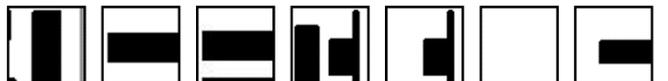

Fig. 6. Sample slices for part-level geometry inputs.

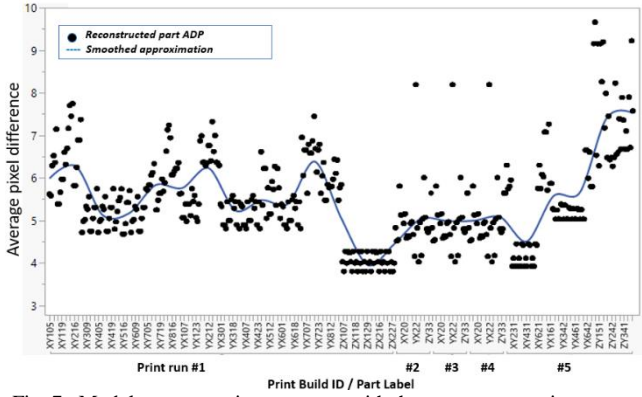
Fig. 7. Model reconstruction accuracy with the part geometry input.

## B. Green part quality prediction

We integrated the well-trained encoder into the subsequent green part quality prediction tasks and demonstrated prediction accuracy by comparing the different model versions. Table. 1. shows the prediction accuracy of TRS bar dimensions measured in the predicted deviation (percentage difference compared to the measured ground-truth value, in mm). The models with thermal input achieve improvements in both prediction accuracy and deviation variation (measured in the standard deviation of deviation distribution) compared to the baseline comparison model without part thermal as input. Our proposed model architecture with encoded part thermal representation achieves the best prediction accuracy. For the predicted error deviation, it shows improvement of 13% (predicted length deviation) to 50% (predicted height deviation) comparing to the thermal sequential input model. We also explored the hyperparameter space of encoded thermal latent vector size with vector sizes of 5, 9, and 20, the preliminary results show no obvious prediction accuracy differences.

## C. Expanding to 3D design data

With a similar network architecture, we replaced the part thermal vector with the corresponding 3D design (slice at each z-height layer, the design data is binary value image with the value of 0 indicates no print region, value of 255 indicates print geometry), and applied a similar data preprocessing pipeline to align the part position inside the print bed and crop the region of interest to form the part geometry 3D vector as input. We experimented with both the reconstruction network and the integrated green quality prediction model. The reasoning for the feasibility of replacing the thermal input with geometry input is that the geometry input contains the factors that are highly correlated to the part thermal distribution during print, including the part geometry, and the part-powder ratio of the neighboring region. We experimented with different ROI for part geometry ROI cropping and adopted the size of 50x50x50 for the below experimental results. Fig. 6

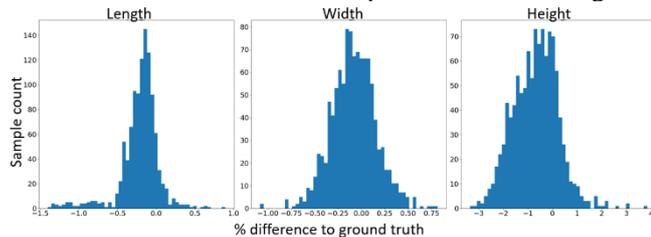
Fig. 8. Predicted green part dimension quality (measured in % deviation).

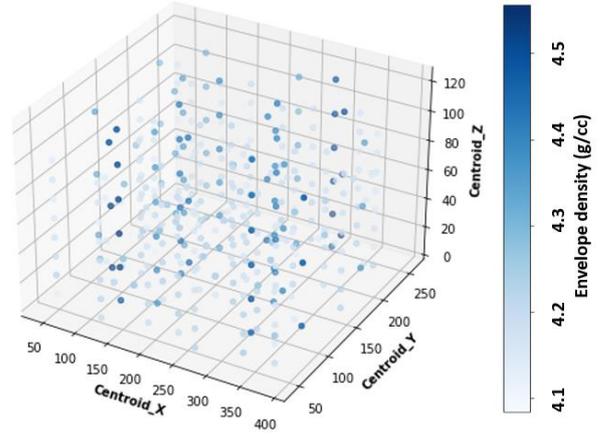
Fig. 9. Predicted green part density value at different location of the print bed.

shows sample slices after pre-processing, taken randomly from the collected builds' design layouts.

Experimental results of part reconstruction are shown in Fig. 7. The reconstructed part shape has an ADP range from 4 to 10. We integrated this well-trained encoder network to the subsequent green quality prediction task and achieved for the best model version, a part dimension accuracy of 0.22% mean difference for length, 0.12% mean difference for width, and 0.48% mean difference for height prediction. This prediction accuracy is comparable to the Table. 1 best model version with multimodal thermal encoder network, demonstrating the possibility of predicting part quality before print, with 3D design and printer parameter instruction as input. Fig. 8 shows the prediction percentage deviation to the ground truth value for the tested TRS parts. Fig. 8 shows the part density prediction results at different locations inside the print bed.

## V. CONCLUSION AND FUTURE WORK

With the proposed model's prediction accuracy and fast inference speed (predict an print bucket's green quality within seconds), this work as a component of HP's Digital Twin effort aims to apply multimodal deep learning to efficiently fuse the data resources from a modality of printer data logs and sensor records, predicting the part quality for a specific printer, a specific print design, a powder batch, a set of print process parameters and the specific part location inside the print bed. Furthermore, this work has demonstrated an outstanding path forward and opened possibilities for many downstream applications to predict and then optimize both the design parameters and the process control parameters, ultimately, the possibility to assist in the improvement of part quality and manufacturing yield.

For future work, we would like to try out different data fusing mechanisms and compare different fusion stages to efficiently interconnect the different modalities' data sources. We could also explore the encoder module architectures, for example, potentially adopt the attention-based architectures to account for the thermal history impact from buried printed layers. We at HP Labs value the significant role of a growing open-source community in accelerating research and development, we welcome boarder collaborations in the field of additive manufacturing and machine learning applications in Digital Twin development.